\documentclass[letterpaper]{article} 
\usepackage{aaai2026}  
\usepackage{times}  
\usepackage{helvet}  
\usepackage{courier}  
\usepackage[hyphens]{url}  
\usepackage{graphicx} 
\usepackage{natbib}  
\usepackage{caption} 
\usepackage{algorithm}
\usepackage{algorithmic}
\usepackage{multirow}
\usepackage{graphicx}
\usepackage{booktabs}     
\usepackage[table]{xcolor}
\usepackage{newfloat}
\usepackage{listings}
\DeclareCaptionStyle{ruled}{labelfont=normalfont,labelsep=colon,strut=off} 
\floatstyle{ruled}
\newfloat{listing}{tb}{lst}{}
\floatname{listing}{Listing}
\title{Empowering Small Language Models with Factual Hallucination-Aware Reasoning for Financial Classification}
\author{
    Han Yuan\equalcontrib,
    Yilin Wu\equalcontrib,
    Li Zhang,
    Zheng Ma\thanks{Correspondence, Singapore Decision Science Center of Excellence, American Express, 1 Marina Boulevard, 018989, Singapore.}
}
\affiliations{
    Global Decision Science, American Express\\
    \texttt{\{Han.Yuan1, yilin.wu, Li.Zhang1, Zheng.Ma2\}@aexp.com}
}

\usepackage{bibentry}

\begin{document}

\maketitle

\begin{abstract}
Small language models (SLMs) are increasingly used for financial classification due to their fast inference and local deployability. However, compared with large language models, SLMs are more prone to factual hallucinations in reasoning and exhibit weaker classification performance. This raises a natural question: \textit{Can mitigating factual hallucinations improve SLMs' financial classification?} To address this, we propose a three-step pipeline named AAAI (\textbf{A}ssociation Identification, \textbf{A}utomated Detection, and \textbf{A}daptive \textbf{I}nference). Experiments on three representative SLMs reveal that: (1) factual hallucinations are positively correlated with misclassifications; (2) encoder-based verifiers effectively detect factual hallucinations; and (3) incorporating feedback on factual errors enables SLMs' adaptive inference that enhances classification performance. We hope this pipeline contributes to trustworthy and effective applications of SLMs in finance.
\end{abstract}


\section{Introduction}
Language models (LMs) are increasingly being deployed for financial classification \citep{guo-etal-2023-chatgpt,li-etal-2023-chatgpt,chen-etal-2024-fintextqa,hu2025extract}. Two main development paths have emerged: one focuses on large language models (LLMs) with superior performance, while the other targets small language models (SLMs) suitable for local deployability \citep{cheng-etal-2024-small}. Although SLMs offer advantages in fast inference and privacy protection, they are prone to factual hallucinations \citep{li-etal-2023-halueval}. A reasoning path containing factual errors undermines both the trustworthiness of an SLM's output and the quality of downstream classification \citep{lin2024towards}. Therefore, enabling SLMs to recognize factual hallucinations in their reasoning potentially enhances the quality of their overall generation.

\begin{figure}[t]
\centering
\includegraphics[width=0.9\columnwidth]{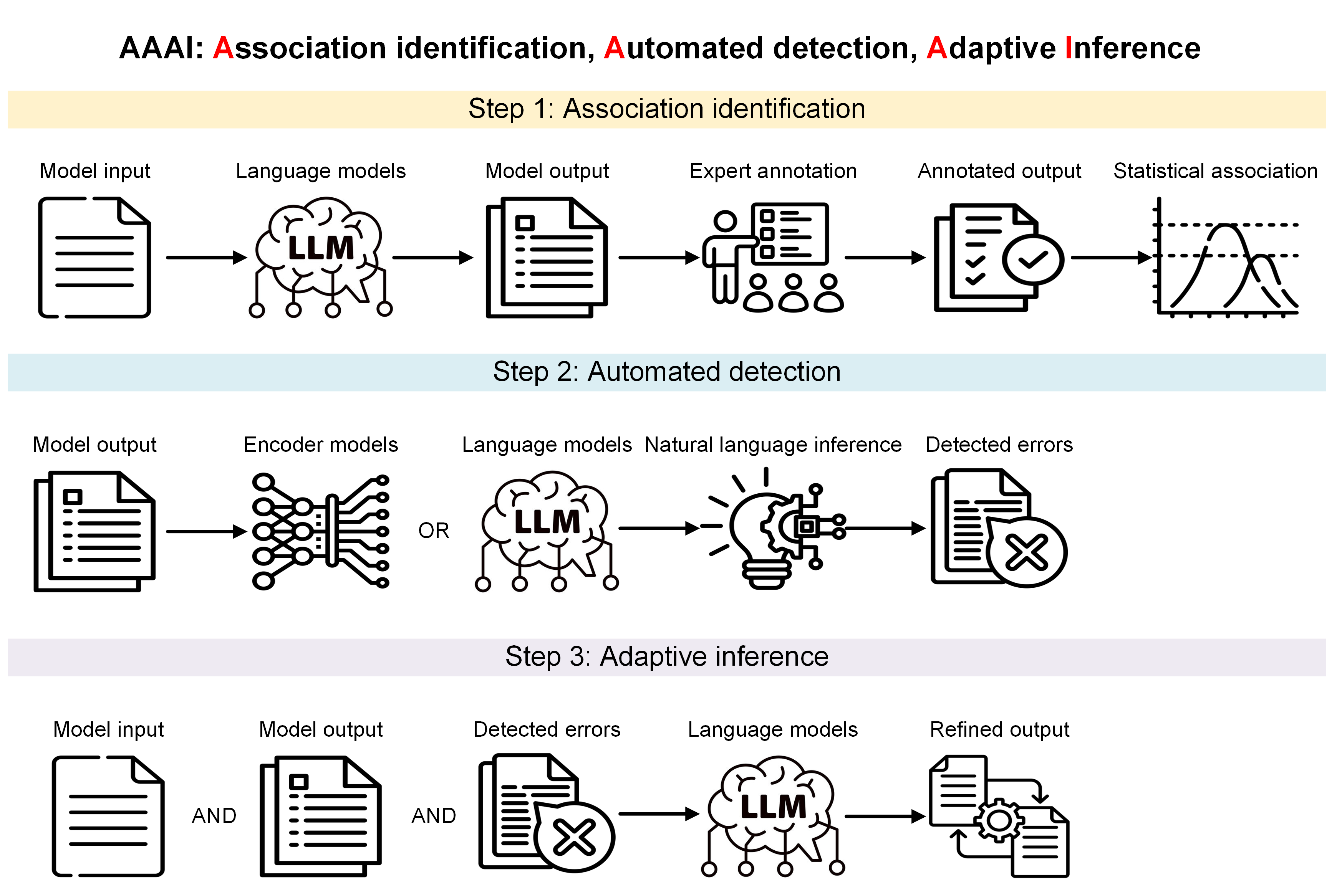}
    \caption{The pipeline for factual error-aware reasoning}
    \label{aaai_pipeline}
\end{figure}

To demonstrate the practical applicability of this approach in finance, we implement a three-step analytical pipeline, abbreviated as AAAI (\textbf{A}ssociation identification, \textbf{A}utomated detection, and \textbf{A}daptive \textbf{I}nference), and Figure~\ref{aaai_pipeline} visualizes each step of this pipeline. First, statistical analysis is conducted to identify the positive association between factual hallucinations in SLMs' reasoning and the accuracy of classifications derived from that reasoning in the initial inference round. This step provides a rationale for improving classification by guiding SLMs to recognize and correct factual errors in their adaptive inference. Second, various methods are used to automatically detect factual errors in the reasoning of the initial inference. This step includes statistical analysis to demonstrate the discriminative ability of the detection models, supporting the scalability of factual error identification. Third, the detected errors are used as feedback for SLMs, prompting adaptive refinement of their inference. Quantitative metrics show that feedback from some methods enhances SLMs' financial classification, while feedback from others degrades it, highlighting the importance of feedback quality in factual error-aware reasoning \citep{huanglarge}.

We emphasize that, unlike studies that use classification correctness to guide SLMs \citep{reflexion,kim2023language}, all feedback in this work focuses on factual issues in LMs' reasoning, which aligns with real-world scenarios where users lack access to the ground truth \citep{lightman2024lets,wang-etal-2024-math}. If the answer were already known, collaboration with LMs would be unnecessary \citep{huanglarge}. However, users can readily identify errors in LMs' reasoning, particularly factual hallucinations that contradict the given context in self-contained analyses where external information is not required \citep{uesato2022,chakraborty2025hallucination}.

\section{Related work}
\subsection{Reasoning and classification}
Recent advancements such as DeepSeek \citep{liu2024deepseek} show the potential of reasoning to enhance model classification performance without explicit user instructions. Although overall performance gains have been observed, some studies further explore the role of reasoning in enhancing generation by LMs \citep{zhao-etal-2023-verify,yuan2025exploring}. \citet{lampinen-etal-2022-language} examine the effect of providing a few in-context examples to LMs' prompts and concluded that explanations improved model performance in general domains. \citet{xi-unreliability} investigate the use of triplets comprising a question, classification, and explanation in few-shot examples, proving LMs tend to generate nonfactual explanations when making wrong predictions. \citet{unfaithful_explanations} also reveal a close relationship between reasoning and decision, showing that even when the decision is incorrect, LMs tend to adjust their explanations to justify it. Our study advances this line by statistically quantifying the relationship between reasoning and classification, providing evidence that factually erroneous reasoning is correlated with misclassification.

\subsection{Factuality verification}
Verifying factual statements and detecting hallucinations are essential for ensuring the trustworthiness and safety of LMs \citep{tang-etal-2024-tofueval,chen-etal-2025-graphcheck}. For LMs' outputs that include both reasoning and classification, hallucination detection can be broadly divided into two types: outcome detection and process detection \citep{welleck2023generating,lightman2024lets}. As discussed previously, outcome detection assumes prior knowledge of whether the outcome is correct, which is often impractical in real-world settings. Former studies have trained process verifiers and demonstrated their effectiveness as feedback to LMs. \citet{lightman2024lets} fine-tune GPT-4 to predict the correctness of each reasoning step in mathematical problems. \citet{wang-etal-2024-math} further extend it by using relatively smaller LMs to provide feedback on each reasoning step and demonstrating that model-generated feedback can guide SLMs to produce better outputs through proximal policy optimization. In this study, we also focus on process verification, specifically examining factual hallucinations within each reasoning step. For backbone models, we adopt transformer encoders for their efficiency and effectiveness as step verifiers of LMs' reasoning \citep{li-etal-2023-making,tang-etal-2023-understanding,sun-etal-2024-towards-verifiable}.

\subsection{Adaptive inference}
Adaptive inference has gained popularity in recent research as a means to enhance LMs' performance through feedback from diverse sources. The previous section reviewed studies that use externally fine-tuned models to verify the factuality of LMs' reasoning, representing one source of feedback. Beyond this, three common sources are knowledge databases, human experts, and other LMs. Feedback from experts serves as the ground truth for assessing the factuality of LMs' reasoning and represents human-in-the-loop practices \citep{yuan2024human}. Knowledge databases are excluded because we focus on self-contained scenarios where all factual context required for correct inference is provided. Feedback from LMs can be categorized into two types: LLM-as-a-judge \citep{jang2022gpt,koutcheme2024open,ye2025learning} and self-reflection \citep{dou-etal-2024-rest,gupta-etal-2024-metareflection,liu2024selectit,zhao-etal-2024-fact,li-etal-2024-think}. Both approaches use LMs to identify errors and guide LMs to iteratively refine earlier outputs. The key difference is that LLM-as-a-judge typically relies on other, often more capable, models, whereas self-reflection uses the same one for initial inference, feedback, and iterative improvement. Compared with general feedback investigated in most former studies, our work focuses specifically on factual hallucinations. Some research also examines adaptive inference based on factuality of LMs' initial outputs. For example, \citet{ji-etal-2023-towards} employ a customized scorer to assess knowledge generated by LMs for open-ended medical tasks. In contrast, our study targets closed-end financial tasks.

\section{Experiments}
\label{experiments_section}
To illustrate the potential for improving SLMs' financial classification through reflection on factual hallucinations, we implement a  three-step AAAI pipeline\footnote{\url{https://github.com/Han-Yuan-Med/Factuality-Aware-Reasoning}} (Figure~\ref{aaai_pipeline}). 

\subsection{Dataset}
\label{dataset_context}
We use the German credit financial classification dataset \citep{german_credit_data}, a widely recognized benchmark in financial natural language processing (NLP). In this context, $L=1$ indicates a good profile and $L=0$ represents a bad profile. Prior studies \cite{pixiu,fintral} on this dataset overlook the critical role of data processing in enhancing the signal-to-noise ratio and revealing the true capabilities of SLMs. Specifically, the original dataset includes outdated information and pre-existing bias \citep{10.1145/3132847.3132938} that pose challenges for SLMs. For instance, certain features are denominated in Deutsche Marks, a currency that has been obsolete for over two decades. Also, SLMs often exhibit limited sensitivity to numeric reasoning \citep{mishra-etal-2022-lila}.
To address these issues, we exclude features misaligned with contemporary SLMs development contexts and converted numeric features into percentile representations. All processing steps were conducted ad hoc and did not involve any operations related to the labels, ensuring that performance was not affected by information leakage. Also, we conduct all experiments using all minority cases paired with an equal number of majority cases to avoid the adverse impact of data imbalance on the analyses.

We utilize three SLMs to generate structured content, containing both reasoning and decision, as the initial responses \citep{yuan2025navigating}: Meta's Llama-3.2-3B \citep{llama}, Google's Gemma-2-2B \cite{gemma}, and Microsoft's Phi-3.5-3.8B \citep{phi}. Following \citet{madaan2023selfrefine}, given an input $X$, prompt $P_{gen}$ and model $M$, an initial generation $Y$ is obtained: $Y=M(P_{gen} \bigoplus X)$. Here, $P_{gen}$ is a task-specific prompt for an initial generation: "Assess the creditworthiness of a customer using the following attributes for financial status. Respond with the final decision of either 'good credit' or 'bad credit' in the first line. Respond with the reasoning on the final decision in the second line. And the attributes are as follows: \{$X$\}. Response: ". $\bigoplus$ stands for concatenation, and $Y$ contains two parts of a classification decision $Y^{cls}$ and $I$ supporting reasoning points $Y_{i}^{rsn}$ $(i=1,2,...,I)$.

For classification metrics, we adopt the standard metrics of F1 score by comparing $Y^{cls}$ with $L$. In addition, financial classification prioritizes weighted costs, emphasizing the greater consequence of false positive to false negative and a lower cost indicates superior performance. As specified in the original dataset documentation \cite{german_credit_data}, the cost associated with a false negative is quantified as 5, while that of a false positive is 1. Therefore, the weighted cost is 
$5\times Num(Y^{cls}=0,L=1)+1\times Num(Y^{cls}=1,L=1)$. We identify a consistent improvement in weighted cost across all models when using the processed dataset. For F1 score, a substantial enhancement is achieved with Llama, while performance remained comparable for the other two.

\begin{table}[]
\centering
\small
\begin{tabular}{ccc}
\hline
SLMs & Coefficient & Risk difference \\ \hline
Llama & 4.47e-2 & 8.75 \\
Gemma & 1.99e-1 & 1.47 \\
Phi & 5.81e-2 & 1.35 \\ \hline
\end{tabular}%
\caption{Positive relationship between factually hallucinated reasoning and misclassifications measured by Pearson correlation and risk difference}
\label{correlation_risk}
\end{table}

\subsection{Association identification}
\label{association_identification}
After evaluating financial classification performance, we conduct an association analysis to examine the relationship between factual hallucinations and misclassifications using Pearson correlation \citep{pearson1895vii}. Basically, for each reasoning point $Y_{i}^{rsn}$, we annotate whether it contains factual hallucinations, where $H_{i}^{rsn}=1$ denotes that $Y_{i}^{rsn}$ contains factual hallucinations. If any $Y_{i}^{rsn}$ contains a factual hallucination, the overall reasoning $Y^{rsn}$ is considered factually erroneous in terms of reasoning ($H^{rsn}=1$). We then examine the correlation between the subgroup of $H^{rsn}=1$ and the subgroup of $Y^{cls} \neq L$. A positive correlation coefficient indicates that reasoning containing factual errors is more likely to occur with incorrect decisions. Also, we calculate the false decision risk difference: $Prob(Y^{cls} \neq L|H^{rsn}=1)-Prob(Y^{cls} \neq L|H^{rsn}=0)$, where $|$ stands for the conditional probability. Similar to the Pearson correlation coefficient, a positive risk difference demonstrates that the risk of misclassification is higher in cases with factual errors than in those without. Table \ref{correlation_risk} presents the correlation results and risk differences, demonstrating the positive relationship between factual hallucinations and incorrect decisions, thereby supporting our subsequent experiments of improving SLMs' classification by mitigating factual errors.

\begin{table}[]
\centering
\small
\begin{tabular}{ccccc}
\hline
SLMs & Verifiers & Mode & AUPRC↑ & BA↑ \\ \hline
\multirow{6}{*}{Llama} & \multirow{2}{*}{DeBERTa} & Pre-trained & 34.04 & 72.66 \\
 &  & FPFT & \cellcolor{blue!15} 82.62 & 80.69 \\
 & \multirow{2}{*}{RoBERTa} & Pre-trained & 55.71 & 74.91 \\
 &  & FPFT & 76.33 & \cellcolor{blue!15} 92.39 \\
 & \multirow{2}{*}{BART} & Pre-trained & 59.72 & 78.36 \\
 &  & FPFT & 76.12 & 83.07 \\ \hline
\multirow{6}{*}{Gemma} & \multirow{2}{*}{DeBERTa} & Pre-trained & 46.44 & 69.98 \\
 &  & FPFT & 96.97 & 96.05 \\
 & \multirow{2}{*}{RoBERTa} & Pre-trained & 25.56 & 59.84 \\
 &  & FPFT & \cellcolor{blue!15} 100.00 & \cellcolor{blue!15} 96.15 \\
 & \multirow{2}{*}{BART} & Pre-trained & 29.19 & 63.36 \\
 &  & FPFT & 90.66 & 93.80 \\ \hline
\multirow{6}{*}{Phi} & \multirow{2}{*}{DeBERTa} & Pre-trained & 26.82 & 58.63 \\
 &  & FPFT & \cellcolor{blue!15} 91.51 & 83.90 \\
 & \multirow{2}{*}{RoBERTa} & Pre-trained & 14.78 & 53.06 \\
 &  & FPFT & 87.29 & \cellcolor{blue!15} 87.39 \\
 & \multirow{2}{*}{BART} & Pre-trained & 22.20 & 56.86 \\
 &  & FPFT & 73.61 & 77.90 \\ \hline
\end{tabular}%
\caption{Performance of verifiers in classifying reasoning points with or without factual hallucinations}
\label{verifier_performance}
\end{table}

\subsection{Automated detection}
\label{automated_detection}
\begin{figure}[t]
\includegraphics[width=\linewidth]{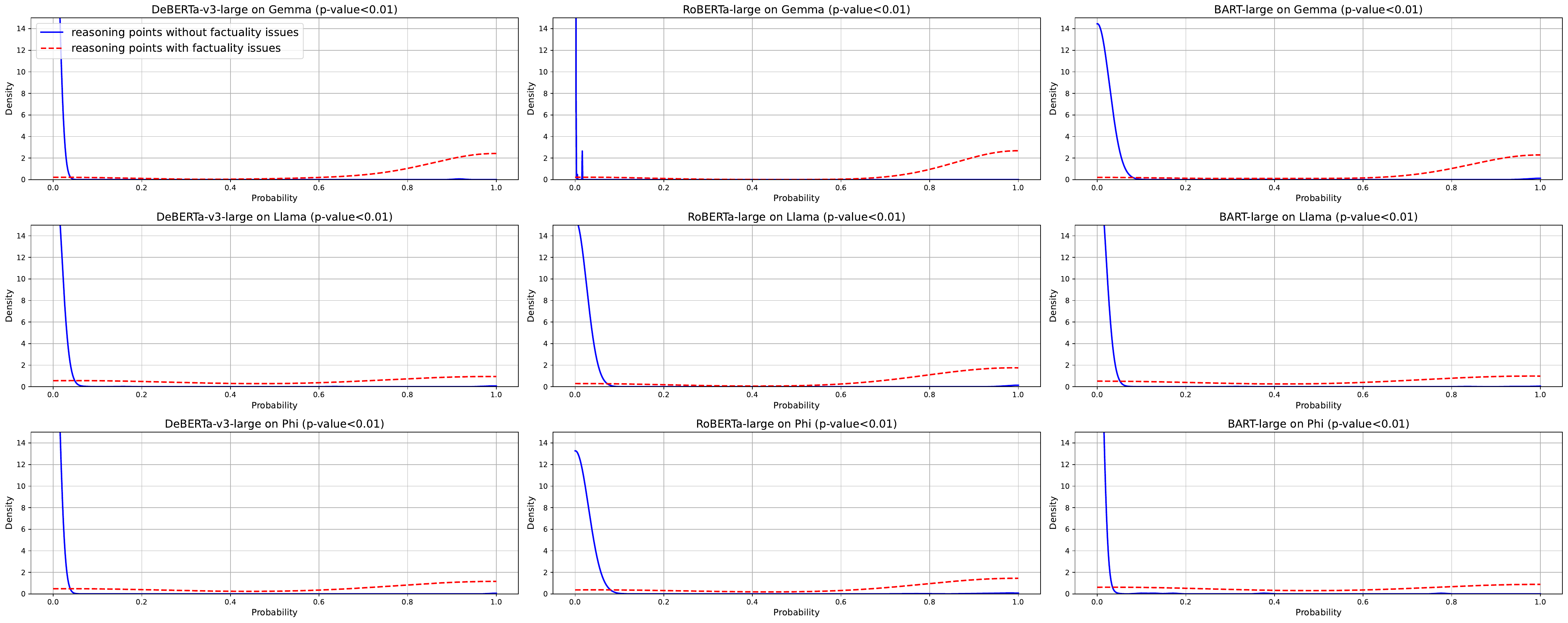}
    \caption{Probability density distribution of fine-tuned verifiers distinguishing reasoning points with and without factual errors}
    \label{pdf}
\end{figure}

Next, we adopt three encoder-based architectures of DeBERTa-v3-large \citep{he2021deberta}, RoBERTa-large \citep{liu2019robertarobustlyoptimizedbert}, and BART-large \citep{lewis2020bart} as verifiers $V$ to predict probability of factual errors $Prob_{i}^{v}=V(X,Y_{i}^{rsn})$ in reasoning steps $Y_{i}^{rsn}$ produced by SLMs \citep{ji-etal-2023-towards,nli-as-a-judge}. To prevent data leakage, full-parameter fine-tuning (FPFT) of $V$ is conducted using a three-fold split, ensuring that $Prob_{i}^{v}$ is collected under the fold where $X$ and $Y_{i}^{rsn}$ serves as test data.
Table \ref{verifier_performance} shows verifiers' performance by comparing $Prob_{i}^{v}$ with $H_{i}^{rsn}$. Due to the limited sample size, we do not employ an independent validation set and instead use a fixed threshold: If $Prob_{i}^{v}\geq 0.5$, the prediction $Pred_{i}^{v}=1$ means that $Y_{i}^{rsn}$ contains factual hallucinations. Also, given the class imbalance, where reasoning points with factual hallucinations are fewer than those without, we evaluate performance using area under the precision-recall curve (AUPRC) and balanced accuracy (BA). In certain cases, a model may achieve a perfect AUPRC score of 1, while the BA remains below 1. In addition to aggregated metrics such as AUPRC, Figure~\ref{pdf} illustrates the probability density distributions of fine-tuned verifiers, showing that they assign substantially higher probabilities to reasoning points with factual errors ($H_{i}^{rsn}=1$) than to those without errors ($H_{i}^{rsn}=0$). We further validate this discriminability using Wilcoxon rank-sum tests \citep{wilcoxon1947probability}, with p-values reported at the top of each subplot. Except for RoBERTa-large on Phi, all p-values are below 0.01, confirming the verifiers' identification ability.



\subsection{Adaptive inference}
\label{adaptive_inference}
Following \citet{kim2023language} and \citet{huanglarge}, we incorporate factual hallucinations in the SLMs' reasoning, detected by diverse methods, as feedback $F_{i}^{rsn}$ to prompt SLMs to refine answers through a tandem round of hallucination-aware reasoning \citep{madaan2023selfrefine}.

Specifically, we employ three strategies to provide feedback and facilitate a tandem round of classification: oracle, verifier, and self-reflection. The oracle provides gold-standard feedback $H_{i}^{rsn}$ from human experts who annotate reasoning steps containing factual errors in the SLMs' initial outputs. The verifier uses the three fine-tuned encoders described in the previous section and outputs the feedback of $Pred_{i}^{v}$. In contrast, self-reflection \citep{renze2024self,multiagent_debate} relies on the same SLMs to detect factual issues in their own reasoning points without leveraging external feedback \citep{huanglarge}. Mathematically, the feedback $Pred_{i}^{m}$ for the reasoning point $Y_{i}^{rsn}$ is $M(P_{fed} \bigoplus X \bigoplus Y_{i}^{rsn})$, where $P_{fed}$ refers to the feedback instruction: "\{$X$\}. Question: does this imply \{$Y_{i}^{rsn}$\}? Yes or No? Response: ".  Based on diverse feedback $F_{i}^{rsn}$ of $H_{i}^{rsn}$, $Pred_{i}^{v}$, and $Pred_{i}^{m}$, the refined generation $Y^{\prime}$ is $M(P_{ref} \bigoplus X \bigoplus Y \bigoplus F_{i}^{rsn})$, where $P_{ref}$ denotes the regeneration instruction based on the feedback: "Your previous response contains the following factual errors: \{$F_{i}^{rsn}$\}. These errors does not match the given attributes. Based on the feedback, improve your decision and reasoning. Response: ". Consistent with the initial generation, $Y^{\prime}$ contains both classification decision $Y^{cls\prime}$ and supporting reasoning, and we evaluate the refined generation $Y^{cls\prime}$ based on $L$.




Table~\ref{reflection_results} compares classification performance across various SLMs and feedback sources. Consistent with \citet{huanglarge} and \citet{madaan2023selfrefine}, our experiments underscore the importance of feedback quality for adaptive inference of SLMs, which can either improve or decline after factual hallucination-aware reasoning. Oracle feedback from human experts consistently enhances, or at least does not reduce, SLMs' performance. Compared with self-reflection, verifiers yield better classification performance in Llama and Gemma, highlighting the need for caution against overreliance on LMs in NLP tasks \citep{tang-etal-2024-tofueval}.

Also, we observe that self-reflection improves Gemma's classification performance, demonstrating the potential of SLMs to correct their own generations without external feedback \citep{wu-etal-2024-large}. Gemma achieves even better classification with feedback from verifiers than from the oracle, indicating that SLMs can produce correct classifications even when feedback is inaccurate, which is reported in \citet{madaan2023selfrefine}. Among these cases, some contain no factual hallucinations in the reasoning, yet classifications are incorrect. A promising study is to explore how multiple-perspective feedback beyond factuality enhances SLMs' reasoning and classification \citep{yan-etal-2024-mirror}.

Moreover, we notice varying levels of steerability across SLMs, where steerability refers to a model’s likelihood of adjusting its output behavior in response to external instructions such as feedback \citep{miehling-etal-2025-evaluating}. Table~\ref{reflection_results} shows that Phi exhibits the lowest steerability, as feedback from oracle, verifiers, or self-reflection does not induce any change from its initial classification, a behavior also observed in Vicuna \cite{madaan2023selfrefine}. Although low steerability limits adjustments to LMs' inherent beliefs, it becomes a desirable property when invalid feedback is given during adaptive inference, as robust LMs should defend their reasoning and decision rather than being easily misled \citep{wang-etal-2023-chatgpt-defend}.

\begin{table}[]
\centering
\small
\begin{tabular}{cccc}
\hline
SLMs & Feedback & F1 score↑ & Weighted cost↓ \\ \hline
\multirow{6}{*}{Llama} & No feedback & 76.42 & 41 \\
 & Oracle & \cellcolor{blue!15} 80.67 & \cellcolor{blue!15} 31 \\
 & Verifier-DeBERTa & 79.66 & 36 \\
 & Verifier-RoBERTa & \cellcolor{blue!15} 80.67 & \cellcolor{blue!15} 31 \\
 & Verifier-BART & 78.99 & 37 \\
 & Self-reflection & 76.42 & 41 \\ \hline
\multirow{6}{*}{Gemma} & No feedback & 67.11 & 49 \\
 & Oracle & 68.49 & 46 \\
 & Verifier-DeBERTa & 68.97 & 45 \\
 & Verifier-RoBERTa & 68.97 & 45 \\
 & Verifier-BART & \cellcolor{blue!15} 69.44 & \cellcolor{blue!15} 44 \\
 & Self-reflection & 67.57 & 48 \\ \hline
\multirow{6}{*}{Phi} & No feedback & 67.11 & 49 \\
 & Oracle & 67.11 & 49 \\
 & Verifier-DeBERTa & 67.11 & 49 \\
 & Verifier-RoBERTa & 67.11 & 49 \\
 & Verifier-BART & 67.11 & 49 \\
 & Self-reflection & 67.11 & 49 \\ \hline
\end{tabular}%
\caption{Performance comparison of SLMs with and without factual hallucination-aware reasoning}
\label{reflection_results}
\end{table}

\section{Supplementary analyses}
In addition to experiments in the former sections, we implement supplementary analyses under alternative settings.

First, we fine-tune SLMs to detect factual errors in their own reasoning and use this feedback to trigger error-aware reasoning. To prevent bias in SLMs’ generation ability, the fine-tuned models are solely used for feedback generation, while the foundation SLMs perform the hallucination-aware reasoning and classification. The resulting classification performance is shown in Table~\ref{slms_finetune}. For Llama, feedback from the fine-tuned model improves classification performance, whereas for the other two SLMs, fine-tuning has marginal effect. Also, compared with feedback from encoder-based verifiers, feedback from either fine-tuned or foundation SLMs does not yield superior performance, indicating that transformer encoders are effective and efficient for initiating SLMs' factual hallucination-aware reasoning.

\begin{table}[]
\centering
\small
\begin{tabular}{cccc}
\hline
SLMs & Feedback & F1 score↑ & Weighted cost↓ \\ \hline
\multirow{2}{*}{Llama} & Fine-tuned & \cellcolor{blue!15} 78.33 & \cellcolor{blue!15} 38 \\
 & Foundation & 76.42 & 41 \\ \hline
\multirow{2}{*}{Gemma} & Fine-tuned & 67.57 & 48 \\
 & Foundation & 67.57 & 48 \\ \hline
\multirow{2}{*}{Phi} & Fine-tuned & 67.11 & 49 \\
 & Foundation & 67.11 & 49 \\ \hline
\end{tabular}%
\caption{Performance comparison of SLMs with feedback from different SLMs' versions}
\label{slms_finetune}
\end{table}

Second, our former experiments adopt a granularity at reasoning point level for self-reflection to maintain comparability with the verifiers. Previous studies \citep{kim2023language,huanglarge} mainly use a coarser granularity, requiring models to reflect on the entire reasoning content containing multiple points. Table~\ref{granularity_table} compares the performance of these two granularities and their impact on SLMs' classification performance. For Llama and Gemma, reasoning at the single point granularity yields better performance across all metrics, likely due to their superior capability on short-context tasks. For Phi, using the entire reasoning content achieves higher F1 score. A possible explanation for this phenomenon is that Phi exhibits lower steerability than the other two SLMs. Since single point granularity provides weaker instructional signals than entire content granularity, Phi's behavior adaptation is consequently less pronounced.

\begin{table}[]
\centering
\small
\begin{tabular}{cccc}
\hline
SLMs & Granularity & F1 score↑ & Weighted cost↓ \\ \hline
\multirow{2}{*}{Llama} & Entire content & 27.27 & 216 \\
 & Single point & \cellcolor{blue!15} 76.42 & \cellcolor{blue!15} 41 \\ \hline
\multirow{2}{*}{Gemma} & Entire content & 22.22 & 221 \\
 & Single point & \cellcolor{blue!15} 67.57 & \cellcolor{blue!15} 48 \\ \hline
\multirow{2}{*}{Phi} & Entire content & \cellcolor{blue!15} 68.70 & 61 \\
 & Single point & 67.11 & \cellcolor{blue!15} 49 \\ \hline
\end{tabular}%
\caption{Performance comparison of SLMs with factual reasoning at different granularities}
\label{granularity_table}
\end{table}

\begin{figure}[t]
\centering
\includegraphics[width=\columnwidth]{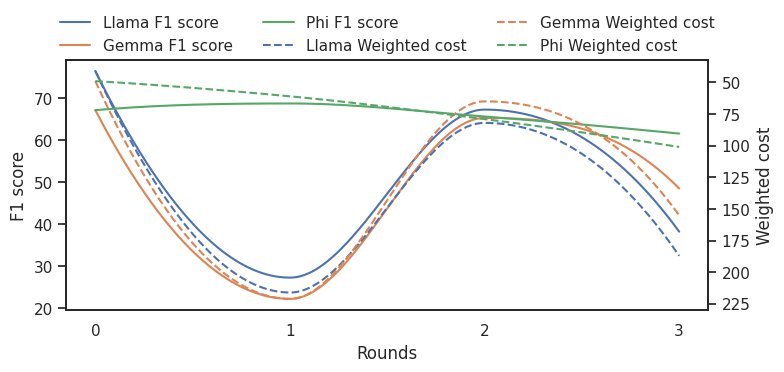}
    \caption{Performance comparison of SLMs across different rounds of reasoning at the entire content granularity}
    \label{several_rounds}
\end{figure}

Third, we explore multiple rounds of self-reflection and adaptive inference. Due to constraints in annotation budget, SLMs' input token length, and GPU memory, we leverage granularity at the entire content level and retain only the latest response and the original context for each additional round of inference. Figure~\ref{several_rounds} shows the impact of additional inference rounds on SLMs' classification performance. Consistent with findings from \citet{huanglarge} in general domain, additional rounds of adaptive inference do not always improve SLMs' performance compared with the initial generation without feedback in financial contexts. Another notable observation is that when a round of hallucination-aware reasoning yields weak performance, the next round often restores it, whereas when a round achieves strong performance, the following round frequently impairs it. Based on the identified positive association between reasoning and classification, we hypothesize that current LMs tend to overcriticize prior reasoning when its quality is high, but provide constructive criticism when its quality is low. The drastic fluctuation across self-reflection rounds presents an opportunity for future foundation model development.


\section{Conclusion and discussion}
We present a three-step approach that allows SLMs to enhance their financial classification by realizing factual errors in their reasoning paths. Compared with prior studies on model reflection, our work introduces statistical analyses to quantify the relationship between erroneous reasoning and misclassifications and to validate the discriminative power of automated detectors. Furthermore, we highlight the importance of pinpointing specific erroneous reasoning steps, which can provide valuable annotation guidance for future SLMs' development \citep{lightman2024lets}.

The positive relationship provides an empirical basis for developing a proxy confidence metric for LMs' classification, such as the proportion of factual errors in the reasoning path. In real-world settings, for example classifying the hawkish or dovish stance in Federal Open Market Committee speeches, even experts often struggle to make accurate decisions. As a result, they cannot always judge whether LMs' classifications are trustworthy, whereas hallucinations of reasoning are easier to evaluate. A proxy confidence value enables users to make more informed decisions about adopting LMs' suggestions. We hope that our study can drive further research on LMs' adaptive inference in finance.

\section*{Limitation}
This work presents a three-step pipeline for studying factual hallucinations in SLMs' reasoning for financial classification and demonstrates the potential to improve SLMs' classification by incorporating factuality into the reasoning. Due to annotation constraints, the experiments are based on 50 positive and 50 negative cases from a public dataset. To validate the generalizability of our findings, future work should include more tasks and SLMs. Beyond the experimental aspect, the technical implementation can also be enhanced. First, the current self-reflection feedback relies on SLMs' zero-shot capability, while providing sufficient few-shot examples may improve their capability. Second, the current work only evaluates factual hallucinations in the initial generation. Although adaptive inference enhances classification, we do not evaluate its impact on reasoning. Further annotation can confirm whether it facilitates self-corrected generation of reasoning sequences \citep{welleck2023generating}.

\section*{Disclaimer}
This paper is provided solely for informational purposes as an academic contribution by the authors to the research community and does not represent, reflect, or constitute the views, policies, positions, or practices of American Express or its affiliates. Nothing in this paper should be cited or relied upon as evidence of, or support for, the business views, policies, positions, or practices of American Express or its affiliates.

\bibliography{aaai2026}

\begin{thebibliography}{58}
\providecommand{\natexlab}[1]{#1}

\bibitem[{Abdin et~al.(2024)Abdin, Aneja, Awadalla et~al.}]{phi}
Abdin, M.; Aneja, J.; Awadalla, H.; et~al. 2024.
\newblock Phi-3 Technical Report: A Highly Capable Language Model Locally on Your Phone.
\newblock \emph{arXiv:2404.14219}.

\bibitem[{Bhatia et~al.(2024)Bhatia, Nagoudi, Cavusoglu et~al.}]{fintral}
Bhatia, G.; Nagoudi, E. M.~B.; Cavusoglu, H.; et~al. 2024.
\newblock {F}in{T}ral: A Family of {GPT}-4 Level Multimodal Financial Large Language Models.
\newblock In \emph{Findings of the Association for Computational Linguistics}.

\bibitem[{Chakraborty, Ornik, and Driggs-Campbell(2025)}]{chakraborty2025hallucination}
Chakraborty, N.; Ornik, M.; and Driggs-Campbell, K. 2025.
\newblock Hallucination detection in foundation models for decision-making: A flexible definition and review of the state of the art.
\newblock \emph{ACM Computing Surveys}, 57(7).

\bibitem[{Chen et~al.(2024)Chen, Zhou, Hua et~al.}]{chen-etal-2024-fintextqa}
Chen, J.; Zhou, P.; Hua, Y.; et~al. 2024.
\newblock {F}in{T}ext{QA}: A Dataset for Long-form Financial Question Answering.
\newblock In \emph{The Annual Meeting of the Association for Computational Linguistics}.

\bibitem[{Chen et~al.(2025)Chen, Liu, Liu, Xie, Yang, Yuan, Fu, Zhou, Chen, Caverlee, and Li}]{chen-etal-2025-graphcheck}
Chen, Y.; Liu, H.; Liu, Y.; Xie, J.; Yang, R.; Yuan, H.; Fu, Y.; Zhou, P.~Y.; Chen, Q.; Caverlee, J.; and Li, I. 2025.
\newblock {G}raph{C}heck: Breaking Long-Term Text Barriers with Extracted Knowledge Graph-Powered Fact-Checking.
\newblock In \emph{Proceedings of the Annual Meeting of the Association for Computational Linguistics}.

\bibitem[{Cheng et~al.(2024)Cheng, Li, Zhao et~al.}]{cheng-etal-2024-small}
Cheng, X.; Li, J.; Zhao, X.; et~al. 2024.
\newblock Small Agent Can Also Rock! Empowering Small Language Models as Hallucination Detector.
\newblock In Al-Onaizan, Y.; Bansal, M.; and Chen, Y.-N., eds., \emph{The Conference on Empirical Methods in Natural Language Processing}.

\bibitem[{Dou et~al.(2024)Dou, Yang, Wu et~al.}]{dou-etal-2024-rest}
Dou, Z.-Y.; Yang, C.-F.; Wu, X.; et~al. 2024.
\newblock Re-{R}e{ST}: Reflection-Reinforced Self-Training for Language Agents.
\newblock In \emph{The Conference on Empirical Methods in Natural Language Processing}.

\bibitem[{Du et~al.(2024)Du, Li, Torralba et~al.}]{multiagent_debate}
Du, Y.; Li, S.; Torralba, A.; et~al. 2024.
\newblock Improving factuality and reasoning in language models through multiagent debate.
\newblock In \emph{The International Conference on Machine Learning}.

\bibitem[{Guo, Xu, and Yang(2023)}]{guo-etal-2023-chatgpt}
Guo, Y.; Xu, Z.; and Yang, Y. 2023.
\newblock Is {C}hat{GPT} a Financial Expert? Evaluating Language Models on Financial Natural Language Processing.
\newblock In \emph{Findings of the Association for Computational Linguistics}.

\bibitem[{Gupta et~al.(2024)Gupta, Kirtania, Singha et~al.}]{gupta-etal-2024-metareflection}
Gupta, P.; Kirtania, S.; Singha, A.; et~al. 2024.
\newblock {M}eta{R}eflection: Learning Instructions for Language Agents using Past Reflections.
\newblock In Al-Onaizan, Y.; Bansal, M.; and Chen, Y.-N., eds., \emph{The Conference on Empirical Methods in Natural Language Processing}.

\bibitem[{He et~al.(2021)He, Liu, Gao et~al.}]{he2021deberta}
He, P.; Liu, X.; Gao, J.; et~al. 2021.
\newblock DeBERTa: Decoding-enhanced BERT with Disentangled Attention.
\newblock In \emph{The International Conference on Learning Representations}.

\bibitem[{Hofmann(1994)}]{german_credit_data}
Hofmann, H. 1994.
\newblock {Statlog (German Credit Data)}.
\newblock UCI Machine Learning Repository.

\bibitem[{Hu et~al.(2025)Hu, Yuan, Pandelea, Luo, Zhao, and Ma}]{hu2025extract}
Hu, B.; Yuan, H.; Pandelea, V.; Luo, W.; Zhao, Y.; and Ma, Z. 2025.
\newblock Extract, Match, and Score: An Evaluation Paradigm for Long Question-context-answer Triplets in Financial Analysis.
\newblock In \emph{CLR 2025 Workshop on Advances in Financial AI: Opportunities, Innovations and Responsible AI}.

\bibitem[{Huang et~al.(2024)Huang, Chen, Mishra et~al.}]{huanglarge}
Huang, J.; Chen, X.; Mishra, S.; et~al. 2024.
\newblock Large Language Models Cannot Self-Correct Reasoning Yet.
\newblock In \emph{The International Conference on Learning Representations}.

\bibitem[{Jang, Lee, and Kim(2022)}]{jang2022gpt}
Jang, Y.; Lee, J.; and Kim, K.-E. 2022.
\newblock GPT-critic: Offline reinforcement learning for end-to-end task-oriented dialogue systems.
\newblock In \emph{The International Conference on Learning Representations}.

\bibitem[{Ji et~al.(2023)Ji, Yu, Xu et~al.}]{ji-etal-2023-towards}
Ji, Z.; Yu, T.; Xu, Y.; et~al. 2023.
\newblock Towards Mitigating {LLM} Hallucination via Self Reflection.
\newblock In \emph{Findings of the Association for Computational Linguistics}.

\bibitem[{Kim, Baldi, and McAleer(2023)}]{kim2023language}
Kim, G.; Baldi, P.; and McAleer, S. 2023.
\newblock Language models can solve computer tasks.
\newblock \emph{The Conference on Neural Information Processing Systems}.

\bibitem[{Koutcheme et~al.(2024)Koutcheme, Dainese, Sarsa et~al.}]{koutcheme2024open}
Koutcheme, C.; Dainese, N.; Sarsa, S.; et~al. 2024.
\newblock Open source language models can provide feedback: Evaluating llms' ability to help students using gpt-4-as-a-judge.
\newblock In \emph{The Innovation and Technology in Computer Science Education}.

\bibitem[{Lampinen et~al.(2022)Lampinen, Dasgupta, Chan et~al.}]{lampinen-etal-2022-language}
Lampinen, A.; Dasgupta, I.; Chan, S.; et~al. 2022.
\newblock Can language models learn from explanations in context?
\newblock In \emph{Findings of the Association for Computational Linguistics}.

\bibitem[{Lewis et~al.(2020)}]{lewis2020bart}
Lewis, M.; et~al. 2020.
\newblock BART: Denoising Sequence-to-Sequence Pre-training for Natural Language Generation, Translation, and Comprehension.
\newblock In \emph{The Annual Meeting of the Association for Computational Linguistics}.

\bibitem[{Li et~al.(2023{\natexlab{a}})Li, Cheng, Zhao et~al.}]{li-etal-2023-halueval}
Li, J.; Cheng, X.; Zhao, X.; et~al. 2023{\natexlab{a}}.
\newblock {H}alu{E}val: A Large-Scale Hallucination Evaluation Benchmark for Large Language Models.
\newblock In \emph{The Conference on Empirical Methods in Natural Language Processing}.

\bibitem[{Li et~al.(2024)Li, Wang, Feng et~al.}]{li-etal-2024-think}
Li, M.; Wang, W.; Feng, F.; et~al. 2024.
\newblock Think Twice Before Trusting: Self-Detection for Large Language Models through Comprehensive Answer Reflection.
\newblock In \emph{Findings of the Association for Computational Linguistics}.

\bibitem[{Li et~al.(2023{\natexlab{b}})Li, Chan, Zhu et~al.}]{li-etal-2023-chatgpt}
Li, X.; Chan, S.; Zhu, X.; et~al. 2023{\natexlab{b}}.
\newblock Are {C}hat{GPT} and {GPT}-4 General-Purpose Solvers for Financial Text Analytics? A Study on Several Typical Tasks.
\newblock In \emph{The Conference on Empirical Methods in Natural Language Processing}.

\bibitem[{Li et~al.(2023{\natexlab{c}})}]{li-etal-2023-making}
Li, Y.; et~al. 2023{\natexlab{c}}.
\newblock Making Language Models Better Reasoners with Step-Aware Verifier.
\newblock In \emph{The Annual Meeting of the Association for Computational Linguistics}.

\bibitem[{Lightman et~al.(2024)Lightman, Kosaraju, Burda et~al.}]{lightman2024lets}
Lightman, H.; Kosaraju, V.; Burda, Y.; et~al. 2024.
\newblock Let's Verify Step by Step.
\newblock In \emph{The International Conference on Learning Representations}.

\bibitem[{Lin et~al.(2024)Lin, Guan, Zhang et~al.}]{lin2024towards}
Lin, Z.; Guan, S.; Zhang, W.; et~al. 2024.
\newblock Towards trustworthy LLMs: a review on debiasing and dehallucinating in large language models.
\newblock \emph{Artificial Intelligence Review}.

\bibitem[{Liu et~al.(2024{\natexlab{a}})Liu, Feng, Wang et~al.}]{liu2024deepseek}
Liu, A.; Feng, B.; Wang, B.; et~al. 2024{\natexlab{a}}.
\newblock Deepseek-v2: A strong, economical, and efficient mixture-of-experts language model.
\newblock \emph{arXiv:2405.04434}.

\bibitem[{Liu et~al.(2024{\natexlab{b}})Liu, Liu, Wong et~al.}]{liu2024selectit}
Liu, L.; Liu, X.; Wong, D.~F.; et~al. 2024{\natexlab{b}}.
\newblock Select{IT}: Selective Instruction Tuning for {LLM}s via Uncertainty-Aware Self-Reflection.
\newblock In \emph{The Conference on Neural Information Processing Systems}.

\bibitem[{Liu et~al.(2019)Liu, Ott, Goyal et~al.}]{liu2019robertarobustlyoptimizedbert}
Liu, Y.; Ott, M.; Goyal, N.; et~al. 2019.
\newblock RoBERTa: A Robustly Optimized BERT Pretraining Approach.
\newblock \emph{arXiv:1907.11692}.

\bibitem[{Madaan et~al.(2023)Madaan, Tandon, Gupta et~al.}]{madaan2023selfrefine}
Madaan, A.; Tandon, N.; Gupta, P.; et~al. 2023.
\newblock Self-Refine: Iterative Refinement with Self-Feedback.
\newblock In \emph{The Conference on Neural Information Processing Systems}.

\bibitem[{Mesnard et~al.(2024)Mesnard, Hardin, Dadashi et~al.}]{gemma}
Mesnard, T.; Hardin, C.; Dadashi, R.; et~al. 2024.
\newblock Gemma: Open Models Based on Gemini Research and Technology.
\newblock \emph{arXiv:2403.08295}.

\bibitem[{Miehling et~al.(2025)Miehling, Desmond, Natesan~Ramamurthy et~al.}]{miehling-etal-2025-evaluating}
Miehling, E.; Desmond, M.; Natesan~Ramamurthy, K.; et~al. 2025.
\newblock Evaluating the Prompt Steerability of Large Language Models.
\newblock In \emph{The Conference of the Nations of the Americas Chapter of the Association for Computational Linguistics}.

\bibitem[{Mishra et~al.(2022)}]{mishra-etal-2022-lila}
Mishra, S.; et~al. 2022.
\newblock {LILA}: A Unified Benchmark for Mathematical Reasoning.
\newblock In \emph{The Conference on Empirical Methods in Natural Language Processing}.

\bibitem[{Pearson(1895)}]{pearson1895vii}
Pearson, K. 1895.
\newblock Note on regression and inheritance in the case of two parents.
\newblock \emph{Proceedings of the Royal Society of London}, 58(347-352): 240--242.

\bibitem[{Renze and Guven(2024)}]{renze2024self}
Renze, M.; and Guven, E. 2024.
\newblock Self-reflection in large language model agents: Effects on problem-solving performance.
\newblock In \emph{The International Conference on Foundation and Large Language Models}.

\bibitem[{Shinn et~al.(2024)Shinn, Cassano, Gopinath et~al.}]{reflexion}
Shinn, N.; Cassano, F.; Gopinath, A.; et~al. 2024.
\newblock Reflexion: language agents with verbal reinforcement learning.
\newblock In \emph{The Conference on Neural Information Processing Systems}.

\bibitem[{Sun et~al.(2024)Sun, Cai, Wang et~al.}]{sun-etal-2024-towards-verifiable}
Sun, H.; Cai, H.; Wang, B.; et~al. 2024.
\newblock Towards Verifiable Text Generation with Evolving Memory and Self-Reflection.
\newblock In \emph{The Conference on Empirical Methods in Natural Language Processing}.

\bibitem[{Tang et~al.(2023)Tang, Goyal, Fabbri et~al.}]{tang-etal-2023-understanding}
Tang, L.; Goyal, T.; Fabbri, A.; et~al. 2023.
\newblock Understanding Factual Errors in Summarization: Errors, Summarizers, Datasets, Error Detectors.
\newblock In \emph{The Annual Meeting of the Association for Computational Linguistics}.

\bibitem[{Tang et~al.(2024)}]{tang-etal-2024-tofueval}
Tang, L.; et~al. 2024.
\newblock {T}ofu{E}val: Evaluating Hallucinations of {LLM}s on Topic-Focused Dialogue Summarization.
\newblock In \emph{The Conference of the North American Chapter of the Association for Computational Linguistics}.

\bibitem[{Touvron et~al.(2023)Touvron, Lavril, Izacard et~al.}]{llama}
Touvron, H.; Lavril, T.; Izacard, G.; et~al. 2023.
\newblock LLaMA: Open and Efficient Foundation Language Models.
\newblock \emph{arXiv:2302.13971}.

\bibitem[{Turpin et~al.(2023)Turpin, Michael, Perez, and Bowman}]{unfaithful_explanations}
Turpin, M.; Michael, J.; Perez, E.; and Bowman, S.~R. 2023.
\newblock Language models don't always say what they think: unfaithful explanations in chain-of-thought prompting.
\newblock In \emph{The Conference on Neural Information Processing Systems}.

\bibitem[{Uesato et~al.(2022)Uesato, Kushman, Kumar et~al.}]{uesato2022}
Uesato, J.; Kushman, N.; Kumar, R.; et~al. 2022.
\newblock Solving math word problems with process-based and outcome-based feedback.
\newblock \emph{arXiv:2211.14275}.

\bibitem[{Wang et~al.(2023)}]{wang-etal-2023-chatgpt-defend}
Wang, B.; et~al. 2023.
\newblock Can {C}hat{GPT} Defend its Belief in Truth? Evaluating {LLM} Reasoning via Debate.
\newblock In \emph{Findings of the Association for Computational Linguistics}.

\bibitem[{Wang et~al.(2024)Wang, Li, Shao et~al.}]{wang-etal-2024-math}
Wang, P.; Li, L.; Shao, Z.; et~al. 2024.
\newblock Math-Shepherd: Verify and Reinforce {LLM}s Step-by-step without Human Annotations.
\newblock In \emph{The Annual Meeting of the Association for Computational Linguistics}.

\bibitem[{Welleck et~al.(2023)Welleck, Lu, West et~al.}]{welleck2023generating}
Welleck, S.; Lu, X.; West, P.; et~al. 2023.
\newblock Generating Sequences by Learning to Self-Correct.
\newblock In \emph{The International Conference on Learning Representations}.

\bibitem[{Wilcoxon(1947)}]{wilcoxon1947probability}
Wilcoxon, F. 1947.
\newblock Probability tables for individual comparisons by ranking methods.
\newblock \emph{Biometrics}, 3(3): 119--122.

\bibitem[{Wu et~al.(2025)Wu, Yuan, Zhang, and Ma}]{nli-as-a-judge}
Wu, Y.; Yuan, H.; Zhang, L.; and Ma, Z. 2025.
\newblock Natural Language Inference as a Judge: Detecting Factuality and Causality Issues in Language Model Self-Reasoning for Financial Analysis.
\newblock In \emph{Proceedings of the Tenth Workshop on Financial Technology and Natural Language Processing}.

\bibitem[{Wu et~al.(2024)}]{wu-etal-2024-large}
Wu, Z.; et~al. 2024.
\newblock Large Language Models Can Self-Correct with Key Condition Verification.
\newblock In \emph{The Conference on Empirical Methods in Natural Language Processing}.

\bibitem[{Xie et~al.(2023)Xie, Han, Zhang et~al.}]{pixiu}
Xie, Q.; Han, W.; Zhang, X.; et~al. 2023.
\newblock PIXIU: A Comprehensive Benchmark, Instruction Dataset and Large Language Model for Finance.
\newblock In \emph{The Conference on Neural Information Processing Systems}.

\bibitem[{Yan et~al.(2024)}]{yan-etal-2024-mirror}
Yan, H.; et~al. 2024.
\newblock Mirror: Multiple-perspective Self-Reflection Method for Knowledge-rich Reasoning.
\newblock In Ku, L.-W.; Martins, A.; and Srikumar, V., eds., \emph{The Annual Meeting of the Association for Computational Linguistics}.

\bibitem[{Ye and Durrett(2024)}]{xi-unreliability}
Ye, X.; and Durrett, G. 2024.
\newblock The unreliability of explanations in few-shot prompting for textual reasoning.
\newblock In \emph{The Conference on Neural Information Processing Systems}.

\bibitem[{Ye et~al.(2025)Ye, Li, Li et~al.}]{ye2025learning}
Ye, Z.; Li, X.; Li, Q.; et~al. 2025.
\newblock Learning LLM-as-a-judge for preference alignment.
\newblock In \emph{The International Conference on Learning Representations}.

\bibitem[{Yuan et~al.(2024)Yuan, Kang, Li, and Fan}]{yuan2024human}
Yuan, H.; Kang, L.; Li, Y.; and Fan, Z. 2024.
\newblock Human-in-the-loop machine learning for healthcare: current progress and future opportunities in electronic health records.
\newblock \emph{Medicine Advances}, 2(3): 318--322.

\bibitem[{Yuan, Zhang, and Ma(2025)}]{yuan2025exploring}
Yuan, H.; Zhang, L.; and Ma, Z. 2025.
\newblock Exploring the Reliability of Self-explanation and its Relationship with Classification in Language Model-driven Financial Analysis.
\newblock In \emph{CLR 2025 Workshop on Advances in Financial AI: Opportunities, Innovations and Responsible AI}.

\bibitem[{Yuan et~al.(2025)Yuan, Zhao, Zhang, Luo, and Ma}]{yuan2025navigating}
Yuan, H.; Zhao, Y.; Zhang, L.; Luo, W.; and Ma, Z. 2025.
\newblock Quantifying the Impact of Structured Output Format on Large Language Models through Causal Inference.
\newblock \emph{arXiv preprint arXiv:2509.21791}.

\bibitem[{Zehlike et~al.(2017)Zehlike, Bonchi, Castillo et~al.}]{10.1145/3132847.3132938}
Zehlike, M.; Bonchi, F.; Castillo, C.; et~al. 2017.
\newblock FA*IR: A Fair Top-k Ranking Algorithm.
\newblock In \emph{The ACM on Conference on Information and Knowledge Management}.

\bibitem[{Zhao et~al.(2023)}]{zhao-etal-2023-verify}
Zhao, R.; et~al. 2023.
\newblock Verify-and-Edit: A Knowledge-Enhanced Chain-of-Thought Framework.
\newblock In \emph{The Annual Meeting of the Association for Computational Linguistics}.

\bibitem[{Zhao et~al.(2024)Zhao, Zhang, Pan et~al.}]{zhao-etal-2024-fact}
Zhao, X.; Zhang, H.; Pan, X.; et~al. 2024.
\newblock Fact-and-Reflection ({F}a{R}) Improves Confidence Calibration of Large Language Models.
\newblock In \emph{Findings of the Association for Computational Linguistics}.

\end{thebibliography}

\end{document}